\documentclass[10pt, a4paper]{article}

\usepackage[final]{lrec-coling2026} 

\usepackage{booktabs}
\usepackage{multirow}
\usepackage{multicol}
\usepackage{fancyvrb}
\usepackage{fvextra}

\title{MazeEval: A Benchmark for Testing Sequential Decision-Making in Language Models}

\name{Hafsteinn Einarsson} 

\address{University of Iceland \\
         Sæmundargata 2, 101 Reykjavík \\
         hafsteinne@hi.is}

\abstract{
As Large Language Models (LLMs) increasingly power autonomous agents in robotics and embodied AI, understanding their spatial reasoning capabilities becomes crucial for ensuring reliable real-world deployment. Despite advances in language understanding, current research lacks evaluation of how LLMs perform spatial navigation without visual cues, a fundamental requirement for agents operating with limited sensory information. This paper addresses this gap by introducing MazeEval, a benchmark designed to isolate and evaluate pure spatial reasoning in LLMs through coordinate-based maze navigation tasks. Our methodology employs a function-calling interface where models navigate mazes of varying complexity ($5\times 5$ to $15\times 15$ grids) using only coordinate feedback and distance-to-wall information, excluding visual input to test fundamental spatial cognition. We evaluate eight state-of-the-art LLMs across identical mazes in both English and Icelandic to assess cross-linguistic transfer of spatial abilities. Our findings reveal striking disparities: while OpenAI's O3 achieves perfect navigation for mazes up to size $30\times 30$, other models exhibit catastrophic failure beyond $9\times 9$ mazes, with 100\% of failures attributed to excessive looping behavior where models revisit a cell at least 10 times. We document a significant performance degradation in Icelandic, with models solving mazes 3-4 sizes smaller than in English, suggesting spatial reasoning in LLMs emerges from linguistic patterns rather than language-agnostic mechanisms. These results have important implications for global deployment of LLM-powered autonomous systems, showing spatial intelligence remains fundamentally constrained by training data availability and highlighting the need for architectural innovations to achieve reliable navigation across linguistic contexts.
 \\ \newline \Keywords{spatial reasoning, agent evaluation, multilingual benchmarks, maze navigation, function calling, sequential decision-making} }

\begin{document}

\maketitleabstract

\section{Introduction}

Spatial reasoning and navigation represent fundamental cognitive abilities that humans employ effortlessly in daily life. As LLMs increasingly serve as the foundation for autonomous agents \cite{duan2022survey}, understanding their capacity for spatial reasoning becomes crucial. While LLMs excel at many language understanding tasks, their ability to maintain spatial awareness and make sequential navigation decisions remains poorly understood~\cite{cohn2023dialectical,sharma2023exploring}.

Recent work has shown that spatial reasoning capabilities did not emerge spontaneously in LLMs the way many other reasoning capabilities did~\cite{li2024advancing}. State-of-the-art models struggled with basic full-information spatial tasks, suffering from consistent performance drops in unfamiliar scenarios \cite{aghzal2023can}. This limitation is a key consideration for deploying LLMs in embodied AI applications that require robust spatial understanding.

We present a comprehensive benchmark designed to evaluate LLMs' spatial reasoning capabilities through maze navigation tasks. Unlike existing benchmarks that often provide visual input or rich environmental descriptions~\cite{chen2024spatialvlm}, our framework challenges models to navigate using only coordinate-based feedback, simulating scenarios where agents must operate with limited sensory information. This constraint reveals fundamental aspects of how LLMs process and reason about spatial relationships.

Importantly, our benchmark represents an \emph{unpolluted} evaluation framework. No dataset of this specific type, namely coordinate-based maze navigation with distance-to-wall feedback, has been publicly released before, ensuring that model providers have not had the opportunity to train their models on similar data motivated by public datasets. This guarantees a fair assessment of genuine spatial reasoning capabilities rather than memorized patterns. The navigation task itself is fairly easy and readers can try it out themselves in an online simulation\footnote{\url{https://haffi112.github.io/maze-navigation-game/}}.

Our central research question is: \emph{How do state-of-the-art LLMs perform on pure spatial reasoning tasks when visual cues are removed and only coordinate-based navigation feedback is available, and does this performance transfer across languages?} This question is particularly relevant as it isolates spatial reasoning from visual processing, tests the fundamental ability to maintain spatial state, and examines whether these capabilities are language-agnostic.

This research has practical implications for the growing use of LLMs in robotics, autonomous navigation, and other embodied AI systems. For these applications, a clear understanding of the models' spatial reasoning limitations is essential for ensuring reliable performance~\cite{firoozi2025foundation,wang2025large,pmlr-v205-ichter23a}. As a recent survey points out, spatial navigation tasks are often deferred to non-LLM algorithms when deploying language models in robotics, which indicates that current models may not be suitable for handling such challenges~\cite{lin2023advances}. Our benchmark's simplicity, using a 2D grid, provides a controlled environment for measuring the fundamental capabilities required in more complex scenarios. Furthermore, the multilingual evaluation (English and Icelandic) offers unique insights into whether spatial reasoning in LLMs is an emergent, language-independent ability or one influenced by the linguistic resources available during training.

The evaluation reveals significant variations in spatial reasoning capabilities across models, with performance degrading substantially as maze size increases. These findings have important implications for deploying LLMs in embodied AI applications and highlight the gap between linguistic and spatial intelligence in current models.

\section{Related Work}

The evaluation of LLMs as autonomous agents has emerged as a critical research area, with several benchmarks addressing different aspects of agent capabilities.

\subsection{General Agent Benchmarks}

Recent work has produced comprehensive benchmarks for evaluating LLM agents. AgentBench~\cite{liu2023agentbench} provides a diverse evaluation suite testing LLMs across web browsing, game playing, and database operations. While comprehensive, it focuses primarily on task completion rather than fundamental reasoning capabilities. WebShop~\cite{yao2022webshop} evaluates web navigation abilities, where agents must navigate e-commerce websites to find and purchase items. While this involves a form of navigation through web pages, it operates in an abstract information space rather than testing physical spatial reasoning or coordinate-based movement.

ALFWorld~\cite{shridhar2021alfworld} and VirtualHome~\cite{puig2018virtualhome} test embodied agents in household environments but assume rich visual or textual descriptions of the environment. Our work differs by focusing on spatial reasoning with minimal sensory input, revealing core navigation capabilities independent of rich environmental descriptions.

\subsection{Spatial Reasoning Benchmarks}

Spatial reasoning in AI has been studied through various lenses. BabyAI~\cite{chevalier2018babyai} provides grid-world navigation tasks but focuses on instruction following rather than autonomous navigation. TextWorld~\cite{cote2018textworld} offers text-based navigation but in richly described environments that provide substantial contextual cues.

The bAbI tasks~\cite{weston2015babi} include spatial reasoning problems but test static spatial relationships rather than dynamic navigation. CLEVR~\cite{johnson2017clevr} evaluates visual reasoning but doesn't address sequential decision-making in navigation contexts. Recent work on the StepGame benchmark~\cite{mirzaee2022stepgame,li2024advancing} has shown that even with dedicated spatial reasoning tasks, LLMs struggle with multi-hop spatial reasoning, particularly when required to track state across multiple steps.

\subsection{Multilingual Evaluation}

While multilingual evaluation has become standard for many NLP tasks~\cite{ahuja2023mega,nielsen-2023-scandeval}, agent benchmarks typically focus on English. Recent work has shown significant performance gaps between English and other languages across various tasks. Lai et al.~\cite{lai2023chatgpt} found a significant drop in ChatGPT's performance on non-English languages across multiple benchmarks. Similarly, evaluations on African languages reveal significant performance gaps~\cite{adelani2024irokosbench}. For code-switching tasks, Zhang et al.~\cite{zhang-etal-2023-multilingual} demonstrated that LLMs show degraded performance compared to monolingual tasks. Our inclusion of Icelandic evaluation provides insights into whether spatial reasoning capabilities transfer across languages, particularly for a morphologically rich language with relatively limited training data compared to English. This addresses a critical gap in understanding whether spatial reasoning in LLMs is language-agnostic or influenced by the linguistic resources available during training.

\section{Methods}

Our evaluation framework consists of three core components: maze generation, LLM interaction interface, and evaluation metrics. We focus on the configuration that provides the most challenging yet fair assessment of spatial reasoning capabilities.

\subsection{Maze Generation}

We generate mazes using a depth-first search (DFS) algorithm, ensuring each maze has exactly one solution path from start to end. Mazes range from $5\times 5$ to $15\times 15$ grids, with complexity increasing not just in size but also in the number of decision points and dead ends. Each maze is characterized by several structural properties that determine its complexity. The basic grid dimensions (width $\times$ height) define the search space, while the optimal path length indicates the minimum number of steps required to reach the goal. We also track the number of decision points, i.e., cells where the agent must choose between more than two open directions since these represent critical junctures where spatial reasoning is most challenged.

\begin{figure}[htbp]
\centering
\includegraphics[width=0.8\columnwidth]{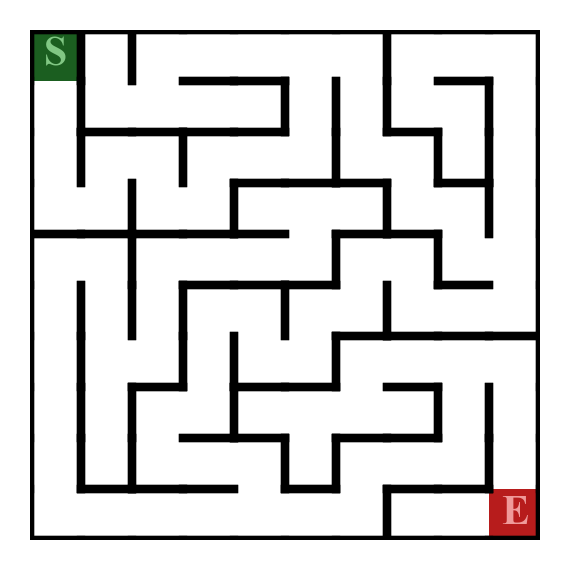}
\caption{Example $10\times 10$ maze showing the start position (S) in green and goal position (E) in red.}
\label{fig:sample_maze}
\end{figure}

For reproducibility, we use fixed seeds for each maze size, generating 5 unique mazes per size category. Figure~\ref{fig:sample_maze} shows an example $10\times 10$ maze used in our evaluation.

\subsection{Decision Information}

Models receive information about how many cells they can move in each direction before hitting a wall. For example, from position (2, 3), the model might see:
\begin{itemize}
\item North: 2 cells (to cell (2, 1))
\item South: 0 cells (wall)
\item East: 3 cells (to cell (5, 3))
\item West: 1 cell (to cell (1, 3))
\end{itemize}

This mode provides more information than binary wall detection but less than a full map view, testing the model's ability to build mental representations from distance information.

Models receive their current position as $(x, y)$ coordinates and the goal position. The coordinate system follows standard computer graphics conventions where the X-axis starts at 0 on the leftmost side and increases rightward (moving east), while the Y-axis begins at 0 at the topmost position and increases downward (moving south). This configuration eliminates visual reasoning entirely, focusing purely on spatial reasoning through numerical coordinates.

\subsection{LLM Interaction Interface}

We implement a function-calling interface where models must invoke a \texttt{move} function with a direction parameter (north, south, east, or west). This approach eliminates parsing ambiguities from natural language responses, ensures consistent action space across all models, and mirrors real-world agent deployments where LLMs must generate executable actions~\cite{schick2023toolformer}. By using function calling, we can precisely measure the model's intended actions without the confounding factor of natural language generation or interpretation errors.

For each step, the model receives four key pieces of information: its current position coordinates in the maze, the distance to walls in each cardinal direction (north, south, east, west), the goal position coordinates it needs to reach, and a complete history of previously visited positions along with the visibility distances that were available at each of those positions. This comprehensive state information allows models to potentially build a mental map of the explored space while making navigation decisions. An example prompt from one of the experiments is shown in Appendix~\ref{ref:prompt}.

To prevent infinite loops and ensure fair comparison, we impose two constraints: each model can visit any cell at most ten times, and the total number of moves is limited to $3n^2$ for an $n \times n$ maze. These limits are generous enough to allow exploration while preventing costly degenerate behaviors.

\subsection{Multilingual Evaluation}

We evaluate models in both English and Icelandic to assess cross-linguistic transfer of spatial reasoning abilities. The Icelandic evaluation uses carefully translated instructions and prompts, with direction names adapted to Icelandic (norður for north, suður for south, austur for east, and vestur for west). Crucially, we use identical maze configurations across both languages to ensure direct comparability of results.

This tests whether spatial reasoning is language-agnostic or influenced by the linguistic framing of the task.

\subsection{Evaluation Metrics}

We track multiple metrics to assess performance. Our primary metrics include the success rate, measuring the percentage of mazes solved within the step limit, and step efficiency, calculated as the ratio of steps taken to the optimal path length. A perfect score would indicate the model found the shortest path, while higher ratios suggest inefficient exploration or backtracking.

Beyond these primary measures, we analyze behavioral patterns that reveal how models approach spatial navigation. We track whether the model failed because it returned to the same cell over ten times or exceeded the movement budget which indicates either poor spatial memory or systematic exploration strategies. We also monitor invalid move attempts, where models try to move through walls, as these reveal fundamental misunderstandings of the spatial constraints or poor processing of the distance feedback.

\subsection{Models Evaluated}

We evaluate a diverse set of state-of-the-art LLMs from the model providers:
\begin{itemize}
\item \textbf{Anthropic}: Claude Sonnet 4 and Claude Opus 4
\item \textbf{Google}: Gemini 2.5 Flash, Gemini 2.5 Pro
\item \textbf{OpenAI}: GPT-4o, GPT-4o-mini, GPT-4.1-mini, O3
\end{itemize}

\subsection{Statistical Analysis}

To rigorously test whether models perform differently between English and Icelandic, we employ the Wilcoxon signed-rank test as our primary statistical method. This non-parametric test is appropriate for our paired data design, where each model solves identical mazes in both languages. The test makes no assumptions about the distribution of performance differences and is robust for small sample sizes.

For each model, we compare success rates across the 11 maze sizes (5×5 to 15×15), treating the success rate at each size as a paired observation. The null hypothesis states that there is no difference in median performance between languages. We use a one-sided test to specifically test whether English performance exceeds Icelandic performance. To account for multiple comparisons across 8 models, we apply Bonferroni correction with an adjusted significance level of $\alpha = 0.05/8 = 0.00625$.

Additionally, we calculate Cohen's d as an effect size measure to quantify the magnitude of performance differences, providing insight beyond statistical significance. For overall comparison across all models, we aggregate success rates to test for a general language effect in spatial reasoning tasks.

\section{Results and Analysis}

We evaluated seven state-of-the-art language models across maze sizes ranging from $5\times 5$ to $15\times 15$ grids in both English and Icelandic. The results reveal significant variations in spatial reasoning capabilities and provide insights into how these abilities transfer across languages.

\subsection{Overall Performance}

\begin{figure}[htbp]
\centering
\includegraphics[width=\columnwidth]{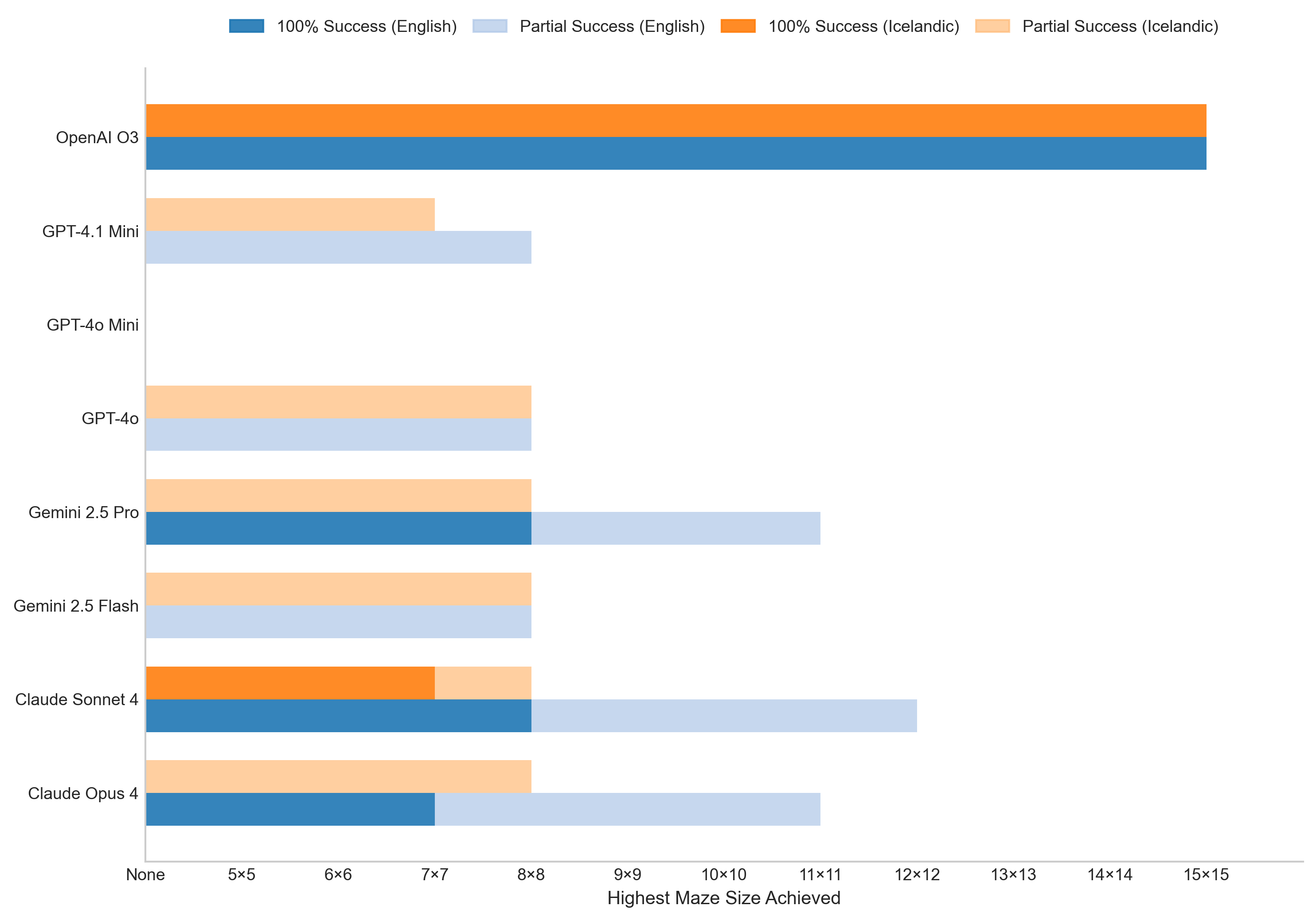}
\caption{Model performance comparison showing the highest difficulty level where each model achieved 100\% success (solid bars) and partial success (translucent bars) for both English and Icelandic evaluations.}
\label{fig:performance_stacked}
\end{figure}

Figure~\ref{fig:performance_stacked} presents the highest difficulty levels achieved by each model. OpenAI's O3 model demonstrated exceptional performance, achieving perfect navigation success across all maze sizes in both languages, making it the only model to do so. In contrast, other models showed substantial performance degradation as maze complexity increased.

\subsection{Performance Degradation with Maze Size}

\begin{figure}[htbp]
\centering
\includegraphics[width=\columnwidth]{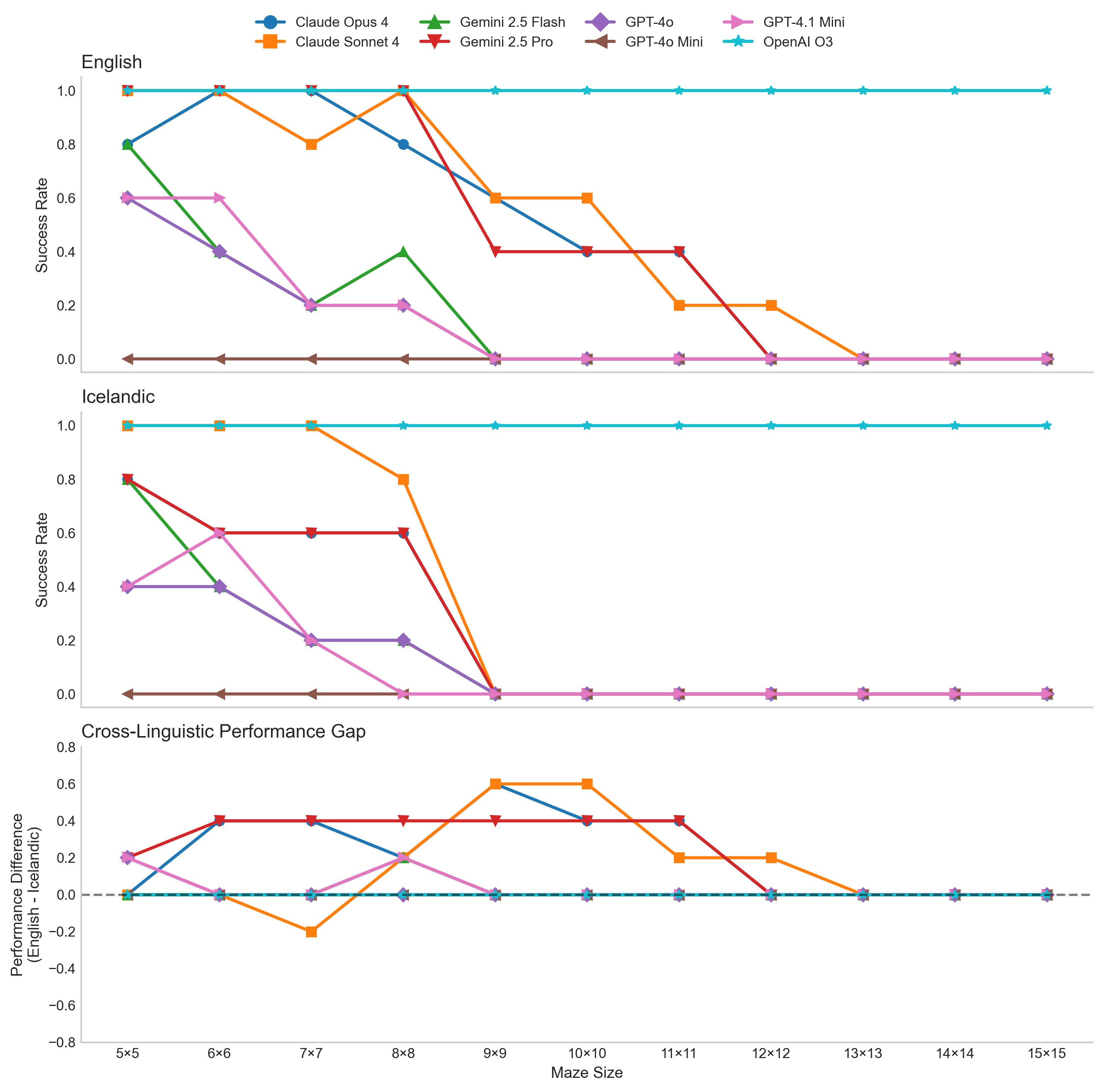}
\caption{Performance of all evaluated models on maze-solving tasks in English and Icelandic. The success rate for most models drops sharply on mazes larger than $7\times 7$. Each model was evaluated on five mazes per size, and testing was stopped at the first size where a model failed all five attempts.}
\label{fig:degradation}
\end{figure}

As shown in Figure~\ref{fig:degradation}, most models exhibit a sharp performance decline as maze size increases. The degradation patterns reveal three distinct performance tiers. At the exceptional level, O3 maintains perfect performance across all maze sizes tested, demonstrating remarkable spatial reasoning capabilities. The strong performers include Claude Opus 4, Claude Sonnet 4, and Gemini 2.5 Pro, which show robust performance up to $8\times 8$ or $9\times 9$ mazes before experiencing significant degradation. The remaining models—GPT-4o, GPT-4o-mini, GPT-4.1-mini, and Gemini 2.5 Flash—struggle beyond $7\times 7$ mazes, with most failing completely at $9\times 9$ and larger sizes.

This tiered performance structure suggests fundamental differences in how models process and maintain spatial information. The sharp transition points, particularly at the $9\times 9$ size threshold, indicate that maze complexity may overwhelm the spatial working memory capacity of most current language models. The consistency of these failure points across multiple runs reinforces that these limitations are systematic rather than stochastic.

\subsection{Cross-Linguistic Analysis}

Contrary to our hypothesis that spatial reasoning would be language-agnostic, we observed notable performance differences between English and Icelandic evaluations. Table~\ref{tab:summary_stats} shows the difference between the two languages.

\begin{table}[htbp]
\centering
\caption{Maximum maze size achieved for English and Icelandic where at least one of the five mazes for the given size was solved.}
\label{tab:summary_stats}
\begin{tabular}{lcc}
\toprule
\multirow{2}{*}{Model} & \multicolumn{2}{c}{Max Maze Size} \\
\cmidrule(lr){2-3}
 & English & Icelandic \\
\midrule
Claude Opus 4 & $11$ & $8$ \\
Claude Sonnet 4 & $12$ & $8$ \\
Gemini 2.5 Flash & $8$ & $8$ \\
Gemini 2.5 Pro & $11$ & $8$ \\
GPT-4o & $8$ & $8$ \\
GPT-4o-mini & - & - \\
GPT-4.1-mini & $8$ & $7$ \\
O3 & $15+$ & $15+$ \\
\bottomrule
\end{tabular}
\end{table}

The analysis of maximum successful maze sizes reveals a consistent pattern of performance degradation in Icelandic compared to English. Most models show a reduction in their maximum solvable maze size when operating in Icelandic, with particularly notable drops for Claude Sonnet 4 (from 12 to 8), Gemini 2.5 Pro (from 11 to 8), and Claude Opus 4 (from 11 to 8). The bottom panel of Figure~\ref{fig:degradation} clearly illustrates this cross-linguistic performance gap, showing positive values (indicating English advantage) for most models across maze sizes. The gap is particularly pronounced for larger mazes, where some models show performance differences of 40-60\% between languages. Only O3 maintains identical performance across both languages, successfully navigating all maze sizes up to $15\times 15$, resulting in a flat line at zero in the difference plot. Due to the cost of evaluation, we did not purse five repeated evaluations of each maze size with O3 further. However, additional experimentation revealed it could solve a maze of size $30\times 30$ for both English and Icelandic and failed to solve a $40\times 40$ maze in both languages.

Our statistical analysis confirms these observations. The Wilcoxon signed-rank test on aggregated data across all models revealed that models performed significantly better in English than Icelandic ($W = 273$, $p < 0.001$, Cohen's $d = 0.50$), representing a medium effect size. Individual model analysis showed that three models exhibited statistically significant differences before correction: Claude Opus 4 ($p = 0.016$, $d = 0.96$), Claude Sonnet 4 ($p = 0.047$, $d = 0.57$), and Gemini 2.5 Pro ($p = 0.008$, $d = 1.20$). However, after applying Bonferroni correction for multiple comparisons ($\alpha = 0.00625$), only the overall comparison remained statistically significant, though the consistent direction of effects across models reinforces the robustness of the language effect and repeated experiments would surely provide the power to detect smaller language performance differences after the correction.

This systematic performance gap suggests that spatial reasoning capabilities in LLMs are indeed influenced by the linguistic resources available during training, with the more limited Icelandic training data potentially constraining the models' ability to process spatial instructions and maintain navigation state.

\subsection{Navigation Efficiency}

\begin{figure}[htbp]
\centering
\includegraphics[width=\columnwidth]{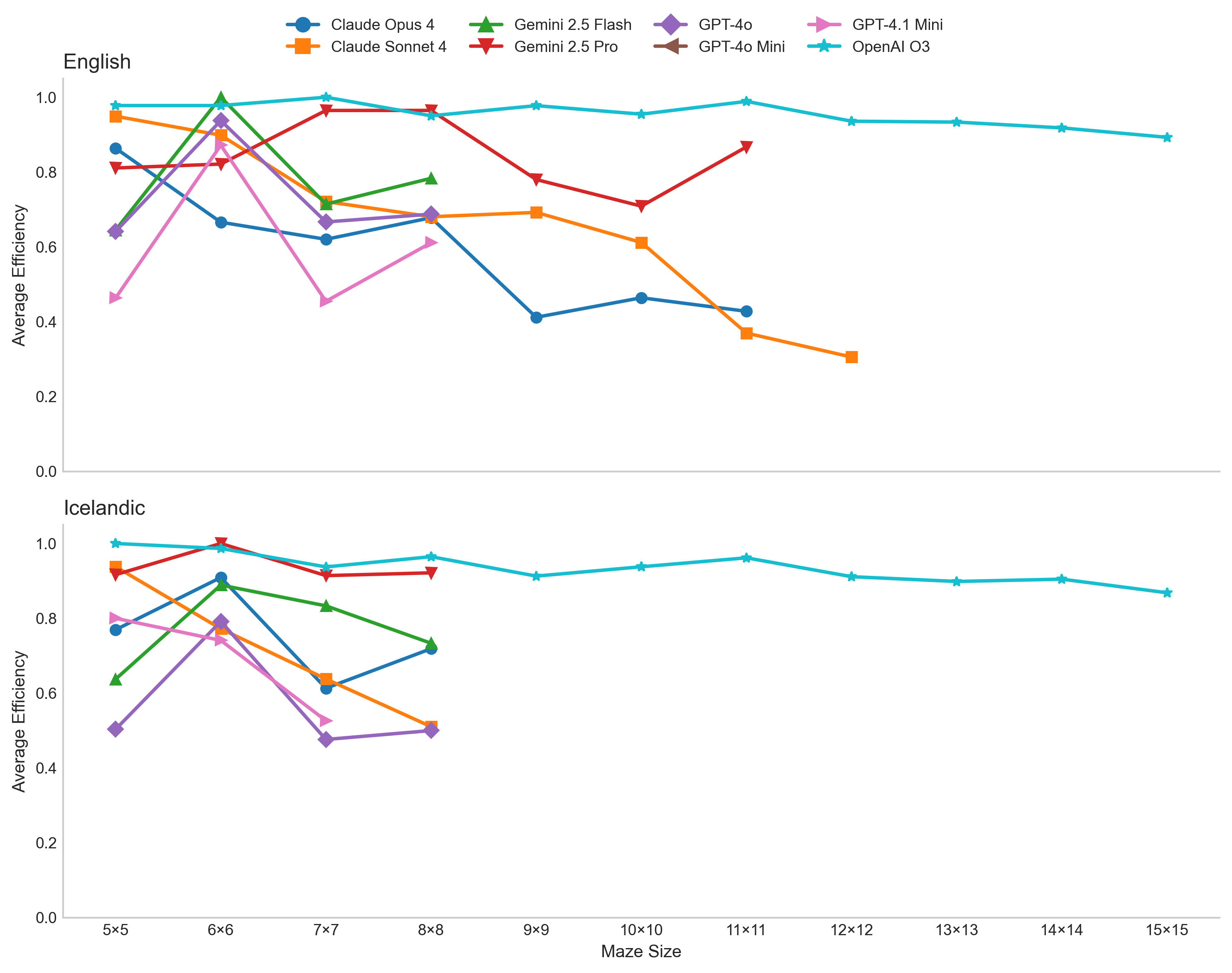}
\caption{Navigation efficiency across maze sizes for successful runs only. Lines show average efficiency (optimal steps / actual steps) with higher values indicating more optimal pathfinding. Note that as maze size increases, these averages represent an increasingly selective subset of mazes that models can still solve.}
\label{fig:efficiency}
\end{figure}

Beyond success rates, we analyzed navigation efficiency, i.e., the ratio of optimal path length to actual steps taken, for all successful runs. Figure~\ref{fig:efficiency} presents the average efficiency across maze sizes from $5\times 5$ to $15\times 15$, calculated only for mazes that models successfully solved. This conditional analysis reveals distinct patterns: O3 maintains near-perfect efficiency (0.89--1.0) across all maze sizes in both languages, demonstrating robust optimal pathfinding capabilities regardless of maze complexity. In contrast, other models show progressive efficiency degradation as maze size increases. Most models achieve reasonable efficiency (0.7--1.0) on smaller mazes but drop to 0.4--0.7 efficiency on larger mazes where they still succeed. Importantly, this efficiency metric is conditioned on success, i.e., as maze size increases, the plotted values represent an increasingly selective subset of ``easier'' mazes that models can still solve. The high variance in efficiency values even among successful runs (e.g., ranging from 0.37 to 1.0 for the same model on similar-sized mazes) suggests that navigation quality depends heavily on specific maze configurations. While the efficiency plots show similar patterns between English and Icelandic for successful runs, the key language difference lies in success rates rather than navigation quality when successful.

\subsection{Failure Analysis}
We analyzed failure cases across all models and languages to understand why models struggle with spatial navigation. Our analysis revealed that 100\% of failures were due to excessive cell visits, where models revisited the same cell 10 or more times before termination. This uniform failure mode reveals a fundamental limitation: models lack the ability to track visited locations despite having access to their navigation history. Notably, no failures occurred due to exceeding the movement budget ($3n^2$ steps for $n\times n$ mazes), indicating that models consistently became trapped in loops well before exhausting their step allocation.

To understand navigation patterns more deeply, we analyzed wall hits and backtracking behavior across all runs (both successful and failed). Detailed visualizations of these metrics are provided in Appendix~\ref{sec:metrics_plots}, showing wall hits and backtracking patterns for each model across all maze sizes. O3 demonstrated exceptional navigation precision with virtually no wall hits and minimal backtracking across all maze sizes in both languages. In contrast, other models showed increasing navigation difficulty with maze complexity. For instance, Claude Opus 4's backtracking increased from 6.8 in $5\times 5$ mazes to 68.6 in $12\times 12$ mazes (English), while Gemini 2.5 Flash exhibited erratic wall collision patterns, spiking to 17.2 average wall hits in $9\times 9$ Icelandic mazes (see Figure~\ref{fig:metrics_gemini_flash} in the appendix). The backtracking analysis reveals that models often explore efficiently in smaller mazes but develop increasingly circular navigation patterns as maze size grows. This pattern tends to be generally worse in Icelandic, suggesting that linguistic context significantly impacts spatial memory maintenance.

\section{Discussion}

Our results reveal a striking dichotomy in spatial reasoning capabilities among current LLMs. While O3 demonstrates that near-perfect maze navigation is achievable, the steep performance degradation observed in other models highlights fundamental challenges in spatial cognition that persist despite advances in language understanding.

\subsection{The Spatial Reasoning Gap}

The dramatic performance drop beyond $7\times 7$ mazes for most models aligns with recent findings that spatial reasoning did not spontaneously emerge in LLMs as other capabilities did~\cite{li2024advancing}. This suggests that current training paradigms, which excel at linguistic tasks, may not adequately develop the spatial representations necessary for navigation. The fact that models struggle with simple 2D grid navigation raises concerns about their deployment in more complex spatial tasks required for embodied AI applications~\cite{lin2023advances}.

The exceptional performance of O3 across all maze sizes indicates that architectural or training innovations can overcome these limitations. This perfect performance across all tested conditions suggests either a fundamental breakthrough in spatial representation learning or targeted optimization for sequential decision-making tasks. The model's ability to maintain coherent navigation state even in $30\times 30$ mazes—where other models fail catastrophically—points to qualitatively different internal representations or reasoning mechanisms. Interestingly, recent work on the ``illusion of thinking''~\cite{illusion-of-thinking} demonstrates that even advanced reasoning models can exhibit systematic failures when problem complexity increases beyond their training distribution. While O3 performs well on mazes up to $30\times 30$ in size, it did exhibit the degradation patterns observed in other models for larger mazes. The illusion-of-thinking framework suggests that apparent reasoning capabilities may mask brittleness when faced with genuinely novel complexity scales. Nevertheless, O3's consistent performance across our benchmark demonstrates that current spatial reasoning limitations in LLMs are not fundamental but rather reflect specific training and architectural choices that can be overcome to some extent.

\subsection{Cross-Linguistic Transfer and Linguistic Relativity}

Our finding that spatial reasoning performance in LLMs systematically degrades in Icelandic compared to English challenges the assumption that such capabilities would be language-agnostic. The consistent reduction in maximum solvable maze sizes, with most models dropping three to four maze size levels, reveals a connection between linguistic resources and spatial reasoning in LLMs. This observation of language influencing a non-linguistic task parallels a long-standing line of inquiry in human cognitive science, which has explored how linguistic structures can affect spatial thinking and navigation strategies~\cite{majid2004can,haun2011plasticity,li2011spatial,levinson1997language}. In a similar vein, recent evaluations of LLMs on agentic tasks have revealed consistent performance gaps across languages, with models showing significant degradation on non-English benchmarks~\cite{lai2023chatgpt,adelani2024irokosbench}, suggesting that linguistic resource availability fundamentally constrains cognitive-like capabilities in artificial systems.

The performance gap likely reflects the vast difference in training data availability between English and Icelandic and a lower emphasis on Icelandic in the posttraining phase of LLM development. With Icelandic having several orders of magnitude less digital text available than English, the limited training corpus appears to prevent models from developing spatial reasoning capabilities on par with English. This suggests that spatial intelligence in LLMs emerges not from explicit spatial reasoning mechanisms but rather through the interaction of linguistic patterns learned across massive text corpora. The finding has important implications for deploying LLMs in multilingual contexts, particularly for languages with limited digital resources, as it reveals that cognitive-like capabilities may be fundamentally constrained by training data availability rather than being universal features of the architecture.

\subsection{Failure Modes and Cognitive Limitations}

The singular failure mode, in which 100\% of failures are caused by excessive cell visits of 10+ visits to the same cell, reveals that models fundamentally lack the ability to integrate information effectively in a spatial memory. Unlike humans who naturally build and update mental maps during navigation~\cite{wang2002human}, most LLMs appear to treat each navigation decision as relatively independent, leading to catastrophic revisiting of previously explored areas. Our analysis of navigation patterns shows that backtracking behavior increases dramatically with maze complexity, with some models averaging over 60 backtracks in larger mazes compared to optimal paths requiring only 20-30 total steps. The models' inability to maintain coherent spatial representations despite having perfect memory of their path history suggests a fundamental limitation in how transformer architectures process and integrate sequential spatial information.

Interestingly, wall collisions were minimal (0.8 hits on average) even in failed runs, indicating that models successfully integrate numerical distance feedback for basic movement constraints.

\subsection{Limitations and Future Directions}

While our benchmark provides valuable insights into LLM spatial reasoning capabilities, several limitations should be acknowledged when interpreting results. The discrete grid representation, though effective for isolating pure spatial reasoning, may not fully capture the continuous nature of real-world navigation where agents must reason about smooth trajectories and variable movement speeds. Our evaluation is limited to 2D mazes, yet many practical applications in robotics and embodied AI require 3D spatial reasoning with additional complexity from vertical navigation and occlusion. The coordinate-based feedback system, while designed to test fundamental capabilities independent of vision, differs from the multimodal inputs typically available in practical applications where visual, tactile, and proprioceptive information complement spatial reasoning. Additionally, our focus on single-agent navigation does not address multi-agent coordination scenarios increasingly important in swarm robotics and collaborative AI systems. Despite these constraints, our benchmark establishes a rigorous baseline for evaluating fundamental spatial reasoning capabilities and reveals systematic limitations that likely extend to more complex spatial tasks.

Future work should explore multiple promising directions to extend and improve upon our findings. First, investigating whether performance improves with visual input or richer environmental descriptions could reveal the role of multimodal information in spatial reasoning. The recent development of visualization-of-thought prompting \cite{vot2024microsoft} offers a particularly promising approach: this technique enables LLMs to generate and reason about visual representations of spatial problems by creating mental imagery through code that produces diagrams or visualizations. By allowing models to \textit{``see''} the spatial layout they are reasoning about, this approach has shown significant improvements in spatial reasoning tasks and might bridge the gap between linguistic and spatial intelligence.

Beyond architectural innovations, context engineering~\cite{liu2023prompting,reynolds2021prompt} presents an alternative pathway for improving navigation performance. Advanced prompting strategies that explicitly scaffold spatial reasoning, such as encouraging models to maintain explicit coordinate maps or systematically plan routes before execution, could potentially overcome some of the limitations we observe. Testing with partial map revelation or memory of previously seen areas could bridge the gap between our minimal feedback and full visibility conditions. Additionally, extending the benchmark to multi-agent navigation scenarios would test higher-level spatial reasoning and coordination capabilities, while moving to three-dimensional mazes or continuous spaces would better approximate real-world navigation challenges. The evaluation could even be extended beyond three dimensions to test truly abstract spatial reasoning capabilities.

\subsection{Towards Spatial Intelligence in LLMs}

Our findings reveal that spatial intelligence remains a fundamental challenge for most LLMs, with O3's perfect performance standing as a notable exception. The universal failure mode of excessive looping suggests that current architectures lack mechanisms analogous to the hippocampal-entorhinal complex that enables biological navigation. In mammalian brains, place cells in the hippocampus and grid cells in the entorhinal cortex form cognitive maps that prevent repetitive exploration through spatial memory consolidation~\cite{matheus2018hippocampal}. Recent work has shown that transformer architectures can learn spatial representations similar to these biological systems when properly configured~\cite{whittington2021relating}, yet most LLMs fail to maintain coherent spatial state during navigation.

The path forward may lie in incorporating neuroscience-inspired architectural innovations. The hippocampus stores and replays spatial trajectories through specialized sequence cells that maintain temporal order~\cite{matheus2018hippocampal}, while grid cells provide a multi-scale coordinate system for efficient spatial coding. Emerging research demonstrates that neural networks trained with brain-inspired learning rules develop grid-like representations spontaneously~\cite{banino2018vector}, suggesting that spatial intelligence could emerge in LLMs through appropriate architectural modifications rather than merely scaling parameters. Such biologically-grounded approaches could address both the catastrophic looping we observe and the language-dependent performance degradation, as spatial memory systems in the brain operate independently of linguistic processing.

The substantial performance gap between English and Icelandic further underscores that current LLMs' spatial reasoning emerges from linguistic patterns rather than dedicated spatial mechanisms. Future architectures might benefit from explicit spatial memory modules inspired by the hippocampal formation, potentially enabling language-agnostic navigation capabilities that mirror the universal nature of spatial cognition in biological systems.

\section{Conclusion}

We presented MazeEval, a comprehensive benchmark for evaluating spatial reasoning and sequential decision-making in Large Language Models through coordinate-based maze navigation. Our evaluation of eight state-of-the-art models across varying maze complexities in both English and Icelandic reveals fundamental limitations in current LLMs' spatial reasoning capabilities, with critical implications for their deployment in embodied AI applications.

Our key findings demonstrate that spatial reasoning remains a significant challenge for most LLMs, with only OpenAI's O3 achieving perfect navigation across all tested conditions. The sharp performance degradation observed in other models at the $9\times 9$ maze threshold, combined with the universal failure mode of excessive looping, suggests that current architectures struggle to maintain coherent spatial representations during sequential navigation tasks. Most significantly, our cross-linguistic analysis reveals a significant performance drop in Icelandic compared to English, with models typically solving mazes 3-4 sizes smaller in the lower-resource language. This systematic degradation extends beyond success rates to navigation efficiency, indicating that linguistic resources fundamentally constrain spatial reasoning capabilities in ways that challenge assumptions about language-agnostic cognitive abilities in AI systems.

These results have important implications for the global deployment of LLM-powered systems in areas like robotics and autonomous navigation, where consistent performance across linguistic contexts is a key factor for safety and reliability. Our benchmark demonstrates that spatial intelligence in current LLMs is not a universal capability but rather a language-dependent skill shaped by training data. As AI systems become more capable, addressing this spatial reasoning gap and ensuring the capability transfers equitably across languages will be a critical step toward developing robust agents that can navigate and interact with the physical world in diverse human environments.

\section{Bibliographical References}\label{sec:reference}

\bibliographystyle{lrec-coling2026-natbib}
\bibliography{references}

\begin{thebibliography}{37}
\expandafter\ifx\csname natexlab\endcsname\relax\def\natexlab#1{#1}\fi

\bibitem[{Adelani et~al.(2025)Adelani, Ojo, Azime, Zhuang, Alabi, He, Ochieng,
  Hooker, Bukula, Lee, Chukwuneke, Buzaaba, Sibanda, Kalipe, Mukiibi,
  Kabongo~Kabenamualu, Yuehgoh, Setaka, Ndolela, Odu, Mabuya, Osei, Muhammad,
  Samb, Guge, Sherman, and Stenetorp}]{adelani2024irokosbench}
David~Ifeoluwa Adelani, Jessica Ojo, Israel~Abebe Azime, Jian~Yun Zhuang,
  Jesujoba~Oluwadara Alabi, Xuanli He, Millicent Ochieng, Sara Hooker, Andiswa
  Bukula, En-Shiun~Annie Lee, Chiamaka~Ijeoma Chukwuneke, Happy Buzaaba,
  Blessing~Kudzaishe Sibanda, Godson~Koffi Kalipe, Jonathan Mukiibi, Salomon
  Kabongo~Kabenamualu, Foutse Yuehgoh, Mmasibidi Setaka, Lolwethu Ndolela,
  Nkiruka Odu, Rooweither Mabuya, Salomey Osei, Shamsuddeen~Hassan Muhammad,
  Sokhar Samb, Tadesse~Kebede Guge, Tombekai~Vangoni Sherman, and Pontus
  Stenetorp. 2025.
\newblock \href {https://doi.org/10.18653/v1/2025.naacl-long.139}
  {{I}roko{B}ench: A new benchmark for {A}frican languages in the age of large
  language models}.
\newblock In \emph{Proceedings of the 2025 Conference of the Nations of the
  Americas Chapter of the Association for Computational Linguistics: Human
  Language Technologies (Volume 1: Long Papers)}, pages 2732--2757,
  Albuquerque, New Mexico. Association for Computational Linguistics.

\bibitem[{Aghzal et~al.(2023)Aghzal, Plaku, and Yao}]{aghzal2023can}
Mohamed Aghzal, Erion Plaku, and Ziyu Yao. 2023.
\newblock Can large language models be good path planners? a benchmark and
  investigation on spatial-temporal reasoning.
\newblock \emph{arXiv preprint arXiv:2310.03249}.

\bibitem[{Ahuja et~al.(2023)Ahuja, Diddee, Hada, Ochieng, Ramesh, Jain, Nambi,
  Ganu, Segal, Ahmed, Bali, and Sitaram}]{ahuja2023mega}
Kabir Ahuja, Harshita Diddee, Rishav Hada, Millicent Ochieng, Krithika Ramesh,
  Prachi Jain, Akshay Nambi, Tanuja Ganu, Sameer Segal, Mohamed Ahmed, Kalika
  Bali, and Sunayana Sitaram. 2023.
\newblock \href {https://doi.org/10.18653/v1/2023.emnlp-main.258} {{MEGA}:
  Multilingual evaluation of generative {AI}}.
\newblock In \emph{Proceedings of the 2023 Conference on Empirical Methods in
  Natural Language Processing}, pages 4232--4267, Singapore. Association for
  Computational Linguistics.

\bibitem[{Banino et~al.(2018)Banino, Barry, Uria, Blundell, Lillicrap,
  Mirowski, Pritzel, Chadwick, Degris, Modayil, Wayne, Soyer, Viola, Zhang,
  Goroshin, Rabinowitz, Pascanu, Beattie, Petersen, Sadik, Gaffney, King,
  Kavukcuoglu, Hassabis, Hadsell, and Kumaran}]{banino2018vector}
Andrea Banino, Caswell Barry, Benigno Uria, Charles Blundell, Timothy
  Lillicrap, Piotr Mirowski, Alexander Pritzel, Martin~J. Chadwick, Thomas
  Degris, Joseph Modayil, Greg Wayne, Hubert Soyer, Fabio Viola, Brian Zhang,
  Ross Goroshin, Neil Rabinowitz, Razvan Pascanu, Charlie Beattie, Stig
  Petersen, Amir Sadik, Stephen Gaffney, Helen King, Koray Kavukcuoglu, Demis
  Hassabis, Raia Hadsell, and Dharshan Kumaran. 2018.
\newblock \href {https://doi.org/10.1038/s41586-018-0102-6} {Vector-based
  navigation using grid-like representations in artificial agents}.
\newblock \emph{Nature}, 557(7705):429--433.

\bibitem[{Chen et~al.(2024)Chen, Xu, Kirmani, Ichter, Sadigh, Guibas, and
  Xia}]{chen2024spatialvlm}
Boyuan Chen, Zhuo Xu, Sean Kirmani, Brain Ichter, Dorsa Sadigh, Leonidas
  Guibas, and Fei Xia. 2024.
\newblock Spatialvlm: Endowing vision-language models with spatial reasoning
  capabilities.
\newblock In \emph{Proceedings of the IEEE/CVF Conference on Computer Vision
  and Pattern Recognition}, pages 14455--14465.

\bibitem[{Chevalier-Boisvert et~al.(2018)Chevalier-Boisvert, Bahdanau, Lahlou,
  Willems, Saharia, Nguyen, and Bengio}]{chevalier2018babyai}
Maxime Chevalier-Boisvert, Dzmitry Bahdanau, Salem Lahlou, Lucas Willems,
  Chitwan Saharia, Thien~Huu Nguyen, and Yoshua Bengio. 2018.
\newblock Babyai: A platform to study the sample efficiency of grounded
  language learning.
\newblock \emph{arXiv preprint arXiv:1810.08272}.

\bibitem[{Cohn and Hernandez-Orallo(2023)}]{cohn2023dialectical}
Anthony~G. Cohn and Jose Hernandez-Orallo. 2023.
\newblock Dialectical language model evaluation: An initial appraisal of the
  commonsense spatial reasoning abilities of llms.
\newblock \emph{arXiv preprint arXiv:2304.11164}.

\bibitem[{C{\^o}t{\'e} et~al.(2018)C{\^o}t{\'e}, K{\'a}d{\'a}r, Yuan, Kybartas,
  Barnes, Fine, Moore, Hausknecht, El~Asri, Adada et~al.}]{cote2018textworld}
Marc-Alexandre C{\^o}t{\'e}, Akos K{\'a}d{\'a}r, Xingdi Yuan, Ben Kybartas,
  Tavian Barnes, Emery Fine, James Moore, Matthew Hausknecht, Layla El~Asri,
  Mahmoud Adada, et~al. 2018.
\newblock Textworld: A learning environment for text-based games.
\newblock In \emph{Workshop on Computer Games}, pages 41--75. Springer.

\bibitem[{Duan et~al.(2022)Duan, Yu, Tan, Zhu, and Tan}]{duan2022survey}
Jiafei Duan, Samson Yu, Hui~Li Tan, Hongyuan Zhu, and Cheston Tan. 2022.
\newblock A survey of embodied ai: From simulators to research tasks.
\newblock \emph{IEEE Transactions on Emerging Topics in Computational
  Intelligence}, 6(2):230--244.

\bibitem[{Firoozi et~al.(2025)Firoozi, Tucker, Tian, Majumdar, Sun, Liu, Zhu,
  Song, Kapoor, Hausman et~al.}]{firoozi2025foundation}
Roya Firoozi, Johnathan Tucker, Stephen Tian, Anirudha Majumdar, Jiankai Sun,
  Weiyu Liu, Yuke Zhu, Shuran Song, Ashish Kapoor, Karol Hausman, et~al. 2025.
\newblock Foundation models in robotics: Applications, challenges, and the
  future.
\newblock \emph{The International Journal of Robotics Research},
  44(5):701--739.

\bibitem[{Haun et~al.(2011)Haun, Rapold, Janzen, and
  Levinson}]{haun2011plasticity}
Daniel~BM Haun, Christian~J Rapold, Gabriele Janzen, and Stephen~C Levinson.
  2011.
\newblock Plasticity of human spatial cognition: Spatial language and cognition
  covary across cultures.
\newblock \emph{Cognition}, 119(1):70--80.

\bibitem[{ichter et~al.(2023)ichter, Brohan, Chebotar, Finn, Hausman, Herzog,
  Ho, Ibarz, Irpan, Jang, Julian, Kalashnikov, Levine, Lu, Parada, Rao,
  Sermanet, Toshev, Vanhoucke, Xia, Xiao, Xu, Yan, Brown, Ahn, Cortes, Sievers,
  Tan, Xu, Reyes, Rettinghouse, Quiambao, Pastor, Luu, Lee, Kuang, Jesmonth,
  Joshi, Jeffrey, Ruano, Hsu, Gopalakrishnan, David, Zeng, and
  Fu}]{pmlr-v205-ichter23a}
brian ichter, Anthony Brohan, Yevgen Chebotar, Chelsea Finn, Karol Hausman,
  Alexander Herzog, Daniel Ho, Julian Ibarz, Alex Irpan, Eric Jang, Ryan
  Julian, Dmitry Kalashnikov, Sergey Levine, Yao Lu, Carolina Parada, Kanishka
  Rao, Pierre Sermanet, Alexander~T Toshev, Vincent Vanhoucke, Fei Xia, Ted
  Xiao, Peng Xu, Mengyuan Yan, Noah Brown, Michael Ahn, Omar Cortes, Nicolas
  Sievers, Clayton Tan, Sichun Xu, Diego Reyes, Jarek Rettinghouse, Jornell
  Quiambao, Peter Pastor, Linda Luu, Kuang-Huei Lee, Yuheng Kuang, Sally
  Jesmonth, Nikhil~J. Joshi, Kyle Jeffrey, Rosario~Jauregui Ruano, Jasmine Hsu,
  Keerthana Gopalakrishnan, Byron David, Andy Zeng, and Chuyuan~Kelly Fu. 2023.
\newblock \href {https://proceedings.mlr.press/v205/ichter23a.html} {Do as i
  can, not as i say: Grounding language in robotic affordances}.
\newblock In \emph{Proceedings of The 6th Conference on Robot Learning}, volume
  205 of \emph{Proceedings of Machine Learning Research}, pages 287--318. PMLR.

\bibitem[{Johnson et~al.(2017)Johnson, Hariharan, Van Der~Maaten, Fei-Fei,
  Lawrence~Zitnick, and Girshick}]{johnson2017clevr}
Justin Johnson, Bharath Hariharan, Laurens Van Der~Maaten, Li~Fei-Fei,
  C~Lawrence~Zitnick, and Ross Girshick. 2017.
\newblock Clevr: A diagnostic dataset for compositional language and elementary
  visual reasoning.
\newblock In \emph{Proceedings of the IEEE conference on computer vision and
  pattern recognition}, pages 2901--2910.

\bibitem[{Lai et~al.(2023)Lai, Ngo, Pouran Ben~Veyseh, Man, Dernoncourt, Bui,
  and Nguyen}]{lai2023chatgpt}
Viet~Dac Lai, Nghia Ngo, Amir Pouran Ben~Veyseh, Hieu Man, Franck Dernoncourt,
  Trung Bui, and Thien~Huu Nguyen. 2023.
\newblock \href {https://doi.org/10.18653/v1/2023.findings-emnlp.878}
  {{C}hat{GPT} beyond {E}nglish: Towards a comprehensive evaluation of large
  language models in multilingual learning}.
\newblock In \emph{Findings of the Association for Computational Linguistics:
  EMNLP 2023}, pages 13171--13189, Singapore. Association for Computational
  Linguistics.

\bibitem[{Levinson(1997)}]{levinson1997language}
Stephen~C Levinson. 1997.
\newblock Language and cognition: The cognitive consequences of spatial
  description in guugu yimithirr.
\newblock \emph{Journal of linguistic anthropology}, 7(1):98--131.

\bibitem[{Li et~al.(2024)Li, Hogg, and Cohn}]{li2024advancing}
Fangjun Li, David~C Hogg, and Anthony~G Cohn. 2024.
\newblock Advancing spatial reasoning in large language models: An in-depth
  evaluation and enhancement using the stepgame benchmark.
\newblock In \emph{Proceedings of the AAAI Conference on Artificial
  Intelligence}, volume~38, pages 18500--18507.

\bibitem[{Li et~al.(2011)Li, Abarbanell, Gleitman, and
  Papafragou}]{li2011spatial}
Peggy Li, Linda Abarbanell, Lila Gleitman, and Anna Papafragou. 2011.
\newblock Spatial reasoning in tenejapan mayans.
\newblock \emph{cognition}, 120(1):33--53.

\bibitem[{Lin et~al.(2023)Lin, Gao, Feng, Xu, Wang, Zhang, Guo, and
  Xu}]{lin2023advances}
Jinzhou Lin, Han Gao, Xuxiang Feng, Rongtao Xu, Changwei Wang, Man Zhang,
  Li~Guo, and Shibiao Xu. 2023.
\newblock Advances in embodied navigation using large language models: A
  survey.
\newblock \emph{arXiv preprint arXiv:2311.00530}.

\bibitem[{Liu et~al.(2023{\natexlab{a}})Liu, Yu, Zhang, Xu, Lei, Lai, Gu, Ding,
  Men, Yang, Zhang, Deng, Zeng, Du, Zhang, Shen, Zhang, Su, Sun, Huang, Dong,
  and Tang}]{liu2023agentbench}
Xiao Liu, Hao Yu, Hanchen Zhang, Yifan Xu, Xuanyu Lei, Hanyu Lai, Yu~Gu,
  Hangliang Ding, Kaiwen Men, Kejuan Yang, Shudan Zhang, Xiang Deng, Aohan
  Zeng, Zhengxiao Du, Chenhui Zhang, Sheng Shen, Tianjun Zhang, Yu~Su, Huan
  Sun, Minlie Huang, Yuxiao Dong, and Jie Tang. 2023{\natexlab{a}}.
\newblock Agentbench: Evaluating llms as agents.
\newblock \emph{arXiv preprint arXiv:2308.03688}.

\bibitem[{Liu et~al.(2023{\natexlab{b}})Liu, Wang, Sun, Yuan, Dong, Di, Wang,
  and Wang}]{liu2023prompting}
Xiaoxia Liu, Jingyi Wang, Jun Sun, Xiaohan Yuan, Guoliang Dong, Peng Di, Wenhai
  Wang, and Dongxia Wang. 2023{\natexlab{b}}.
\newblock Prompting frameworks for large language models: A survey.
\newblock \emph{arXiv preprint arXiv:2311.12785}.

\bibitem[{Majid et~al.(2004)Majid, Bowerman, Kita, Haun, and
  Levinson}]{majid2004can}
Asifa Majid, Melissa Bowerman, Sotaro Kita, Daniel~BM Haun, and Stephen~C
  Levinson. 2004.
\newblock Can language restructure cognition? the case for space.
\newblock \emph{Trends in cognitive sciences}, 8(3):108--114.

\bibitem[{Matheus~Gauy et~al.(2018)Matheus~Gauy, Lengler, Einarsson, Meier,
  Weissenberger, Yanik, and Steger}]{matheus2018hippocampal}
Marcelo Matheus~Gauy, Johannes Lengler, Hafsteinn Einarsson, Florian Meier,
  Felix Weissenberger, Mehmet~Fatih Yanik, and Angelika Steger. 2018.
\newblock A hippocampal model for behavioral time acquisition and fast
  bidirectional replay of spatio-temporal memory sequences.
\newblock \emph{Frontiers in neuroscience}, 12:961.

\bibitem[{Mirzaee and Kordjamshidi(2022)}]{mirzaee2022stepgame}
Roshanak Mirzaee and Parisa Kordjamshidi. 2022.
\newblock Transfer learning with synthetic corpora for spatial role labeling
  and reasoning.
\newblock In \emph{Proceedings of the 2022 Conference on Empirical Methods in
  Natural Language Processing}, pages 6148--6165.

\bibitem[{Nielsen(2023)}]{nielsen-2023-scandeval}
Dan Nielsen. 2023.
\newblock \href {https://aclanthology.org/2023.nodalida-1.20/} {{S}cand{E}val:
  A benchmark for {S}candinavian natural language processing}.
\newblock In \emph{Proceedings of the 24th Nordic Conference on Computational
  Linguistics (NoDaLiDa)}, pages 185--201, T{\'o}rshavn, Faroe Islands.
  University of Tartu Library.

\bibitem[{Puig et~al.(2018)Puig, Ra, Boben, Li, Wang, Fidler, and
  Torralba}]{puig2018virtualhome}
Xavier Puig, Kevin Ra, Marko Boben, Jiaman Li, Tingwu Wang, Sanja Fidler, and
  Antonio Torralba. 2018.
\newblock Virtualhome: Simulating household activities via programs.
\newblock In \emph{Proceedings of the IEEE conference on computer vision and
  pattern recognition}, pages 8494--8502.

\bibitem[{Reynolds and McDonell(2021)}]{reynolds2021prompt}
Laria Reynolds and Kyle McDonell. 2021.
\newblock Prompt programming for large language models: Beyond the few-shot
  paradigm.
\newblock In \emph{Extended abstracts of the 2021 CHI conference on human
  factors in computing systems}, pages 1--7.

\bibitem[{Schick et~al.(2023)Schick, Dwivedi-Yu, Dess{\`\i}, Raileanu, Lomeli,
  Hambro, Zettlemoyer, Cancedda, and Scialom}]{schick2023toolformer}
Timo Schick, Jane Dwivedi-Yu, Roberto Dess{\`\i}, Roberta Raileanu, Maria
  Lomeli, Eric Hambro, Luke Zettlemoyer, Nicola Cancedda, and Thomas Scialom.
  2023.
\newblock Toolformer: Language models can teach themselves to use tools.
\newblock \emph{Advances in Neural Information Processing Systems},
  36:68539--68551.

\bibitem[{Sharma(2023)}]{sharma2023exploring}
Manasi Sharma. 2023.
\newblock Exploring and improving the spatial reasoning abilities of large
  language models.
\newblock \emph{arXiv preprint arXiv:2312.01054}.

\bibitem[{Shojaee*† et~al.(2025)Shojaee*†, Mirzadeh*, Alizadeh, Horton,
  Bengio, and Farajtabar}]{illusion-of-thinking}
Parshin Shojaee*†, Iman Mirzadeh*, Keivan Alizadeh, Maxwell Horton, Samy
  Bengio, and Mehrdad Farajtabar. 2025.
\newblock \href
  {https://ml-site.cdn-apple.com/papers/the-illusion-of-thinking.pdf} {The
  illusion of thinking: Understanding the strengths and limitations of
  reasoning models via the lens of problem complexity}.

\bibitem[{Shridhar et~al.(2021)Shridhar, Yuan, C\^ot\'e, Bisk, Trischler, and
  Hausknecht}]{shridhar2021alfworld}
Mohit Shridhar, Xingdi Yuan, Marc-Alexandre C\^ot\'e, Yonatan Bisk, Adam
  Trischler, and Matthew Hausknecht. 2021.
\newblock \href {https://arxiv.org/abs/2010.03768} {{ALFWorld: Aligning Text
  and Embodied Environments for Interactive Learning}}.
\newblock In \emph{Proceedings of the International Conference on Learning
  Representations (ICLR)}.

\bibitem[{Wang et~al.(2025)Wang, Shi, Hu, Ma, Liu, Wang, Yao, Liu, Ge, and
  Zhang}]{wang2025large}
Jiaqi Wang, Enze Shi, Huawen Hu, Chong Ma, Yiheng Liu, Xuhui Wang, Yincheng
  Yao, Xuan Liu, Bao Ge, and Shu Zhang. 2025.
\newblock Large language models for robotics: Opportunities, challenges, and
  perspectives.
\newblock \emph{Journal of Automation and Intelligence}, 4(1):52--64.

\bibitem[{Wang and Spelke(2002)}]{wang2002human}
Ranxiao~Frances Wang and Elizabeth~S Spelke. 2002.
\newblock Human spatial representation: Insights from animals.
\newblock \emph{Trends in cognitive sciences}, 6(9):376--382.

\bibitem[{Weston et~al.(2015)Weston, Bordes, Chopra, Rush, Van~Merri{\"e}nboer,
  Joulin, and Mikolov}]{weston2015babi}
Jason Weston, Antoine Bordes, Sumit Chopra, Alexander~M Rush, Bart
  Van~Merri{\"e}nboer, Armand Joulin, and Tomas Mikolov. 2015.
\newblock Towards ai-complete question answering: A set of prerequisite toy
  tasks.
\newblock \emph{arXiv preprint arXiv:1502.05698}.

\bibitem[{Whittington et~al.(2021)Whittington, Warren, and
  Behrens}]{whittington2021relating}
James~CR Whittington, Joseph Warren, and Timothy~EJ Behrens. 2021.
\newblock Relating transformers to models and neural representations of the
  hippocampal formation.
\newblock \emph{arXiv preprint arXiv:2112.04035}.

\bibitem[{Wu et~al.(2024)Wu, Mao, Zhang, Xia, Dong, Cui, and
  Wei}]{vot2024microsoft}
Wenshan Wu, Shaoguang Mao, Yadong Zhang, Yan Xia, Li~Dong, Lei Cui, and Furu
  Wei. 2024.
\newblock Mind's eye of llms: visualization-of-thought elicits spatial
  reasoning in large language models.
\newblock \emph{Advances in Neural Information Processing Systems},
  37:90277--90317.

\bibitem[{Yao et~al.(2022)Yao, Chen, Yang, and Narasimhan}]{yao2022webshop}
Shunyu Yao, Howard Chen, John Yang, and Karthik Narasimhan. 2022.
\newblock Webshop: Towards scalable real-world web interaction with grounded
  language agents.
\newblock \emph{Advances in Neural Information Processing Systems},
  35:20744--20757.

\bibitem[{Zhang et~al.(2023)Zhang, Cahyawijaya, Cruz, Winata, and
  Aji}]{zhang-etal-2023-multilingual}
Ruochen Zhang, Samuel Cahyawijaya, Jan Christian~Blaise Cruz, Genta Winata, and
  Alham~Fikri Aji. 2023.
\newblock \href {https://doi.org/10.18653/v1/2023.emnlp-main.774} {Multilingual
  large language models are not (yet) code-switchers}.
\newblock In \emph{Proceedings of the 2023 Conference on Empirical Methods in
  Natural Language Processing}, pages 12567--12582, Singapore. Association for
  Computational Linguistics.

\end{thebibliography}


\appendix

\section{Prompt}\label{ref:prompt}
The following is an example of a prompt for a $7\times 7$ maze where an agent has already performed several move actions.
\begin{Verbatim}[breaklines=true, breakanywhere=true]
You are navigating a maze. Your goal is to reach the end position.

The coordinate system works as follows:
- X-axis (first number): 0 is leftmost, higher values move right (east)
- Y-axis (second number): 0 is topmost, higher values move down (south)
- North decreases Y, south increases Y, east increases X, west decreases X

You can see how many cells you can move in each direction:
- North: 2 cells (to cell (0, 2))
- South: 3 cells (to cell (0, 4))
- East: 0 cells
- West: 0 cells

Your current position is (0, 3).
The goal is at position (6, 6).

Your travel history (with what you observed at each position):
Note: Numbers show how many cells you could see in each direction (e.g., 2E means 2 cells east)
1. At (0, 0) [saw: 1E, 0W, 0S, 0N] - Moved east to position (1, 0)
2. At (1, 0) [saw: 0E, 1W, 2S, 0N] - Moved south to position (1, 1)
3. At (1, 1) [saw: 0E, 0W, 1S, 1N] - Moved south to position (1, 2)
4. At (1, 2) [saw: 0E, 1W, 0S, 2N] - Moved west to position (0, 2)
5. At (0, 2) [saw: 1E, 0W, 4S, 1N] - Moved south to position (0, 3)

Use the move function to navigate. Choose a direction: north, south, east, or west.

WARNING: If you visit the same cell 10 times, the evaluation will be terminated. Avoid getting stuck in loops!
\end{Verbatim}

\section{Navigation Metrics Analysis}\label{sec:metrics_plots}

This appendix presents detailed visualizations of wall hits and backtracking behavior for each model across all maze sizes in both English and Icelandic evaluations. These metrics provide deeper insights into the navigation patterns that lead to success or failure.

The figures below show two key metrics:
\begin{itemize}
\item \textbf{Wall Hits}: Average number of attempts to move through walls per maze, indicating spatial awareness
\item \textbf{Backtracks}: Average number of times the model returns to previously visited cells, indicating exploration efficiency
\end{itemize}

\begin{figure}[htbp]
\centering
\includegraphics[width=\columnwidth]{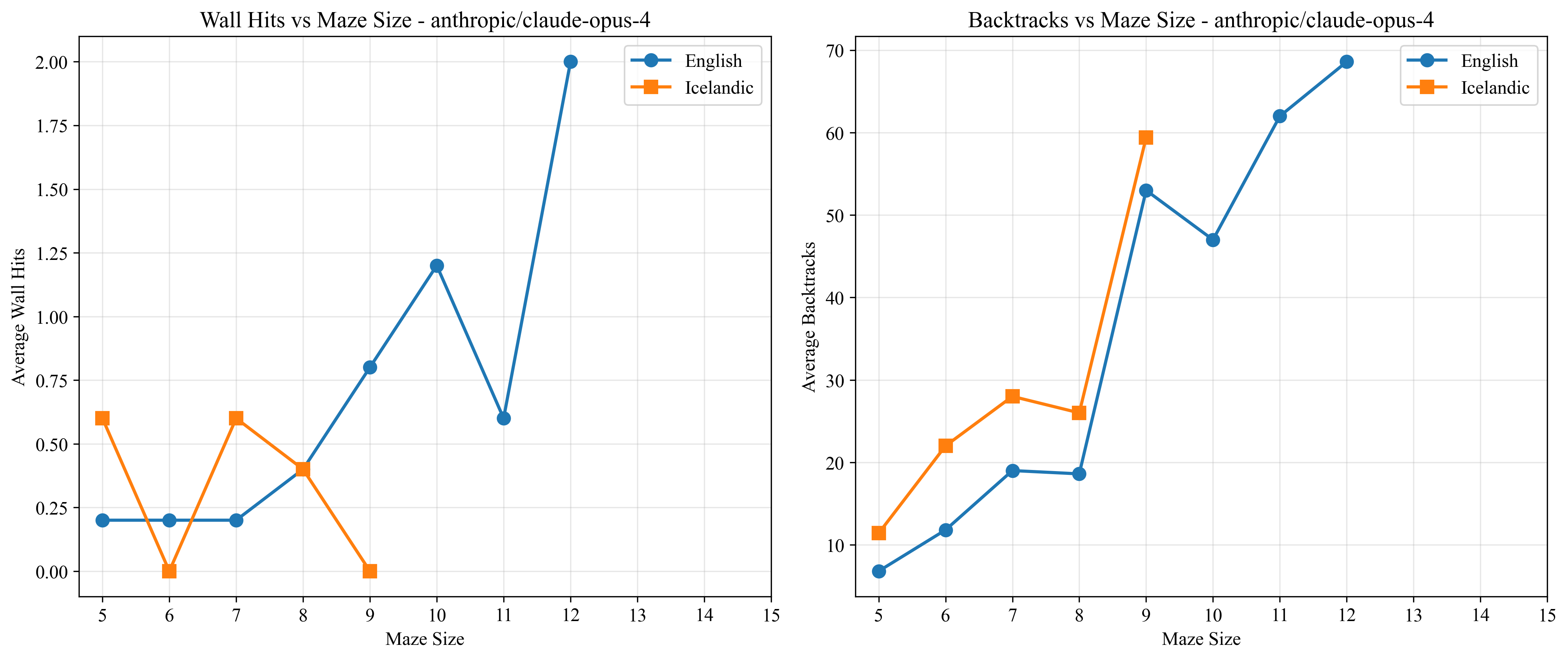}
\caption{Wall hits and backtracking patterns for Claude Opus 4 across maze sizes.}
\label{fig:metrics_claude_opus}
\end{figure}

\begin{figure}[htbp]
\centering
\includegraphics[width=\columnwidth]{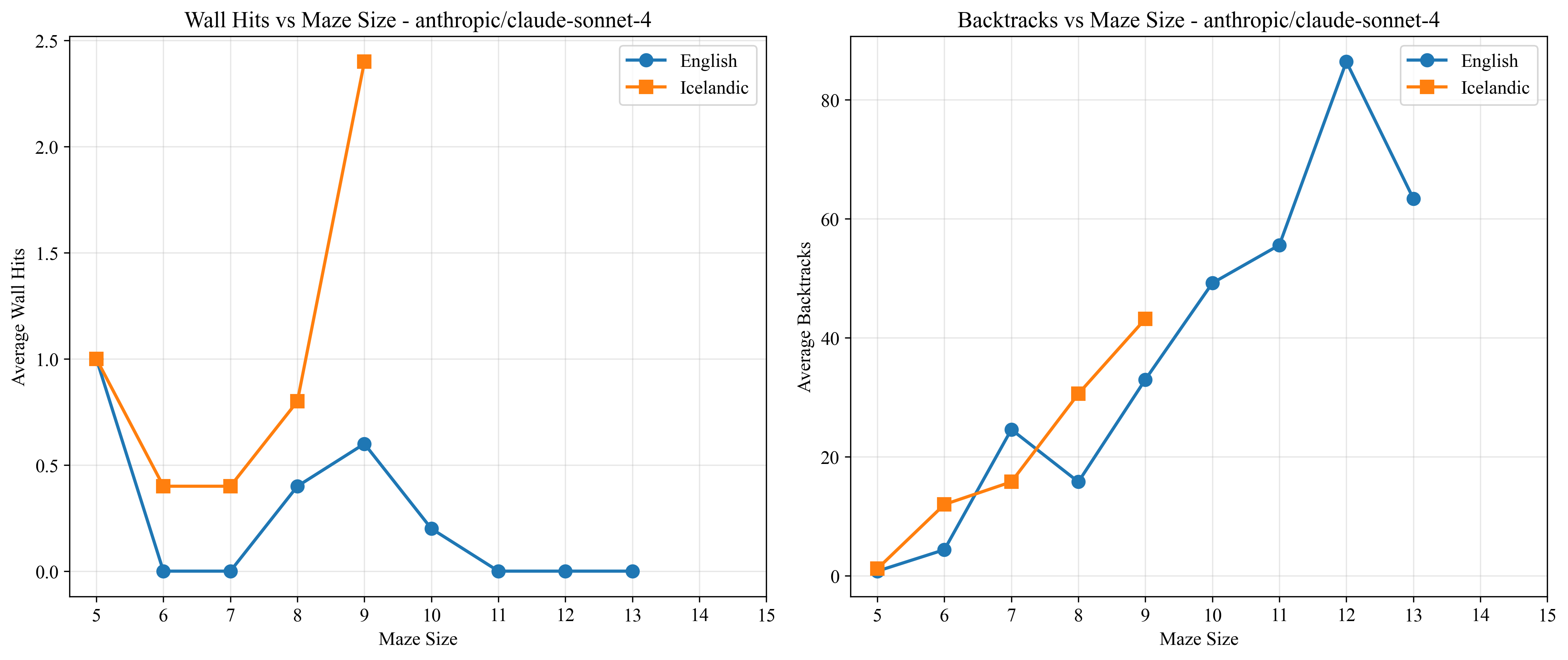}
\caption{Wall hits and backtracking patterns for Claude Sonnet 4 across maze sizes.}
\label{fig:metrics_claude_sonnet}
\end{figure}

\begin{figure}[htbp]
\centering
\includegraphics[width=\columnwidth]{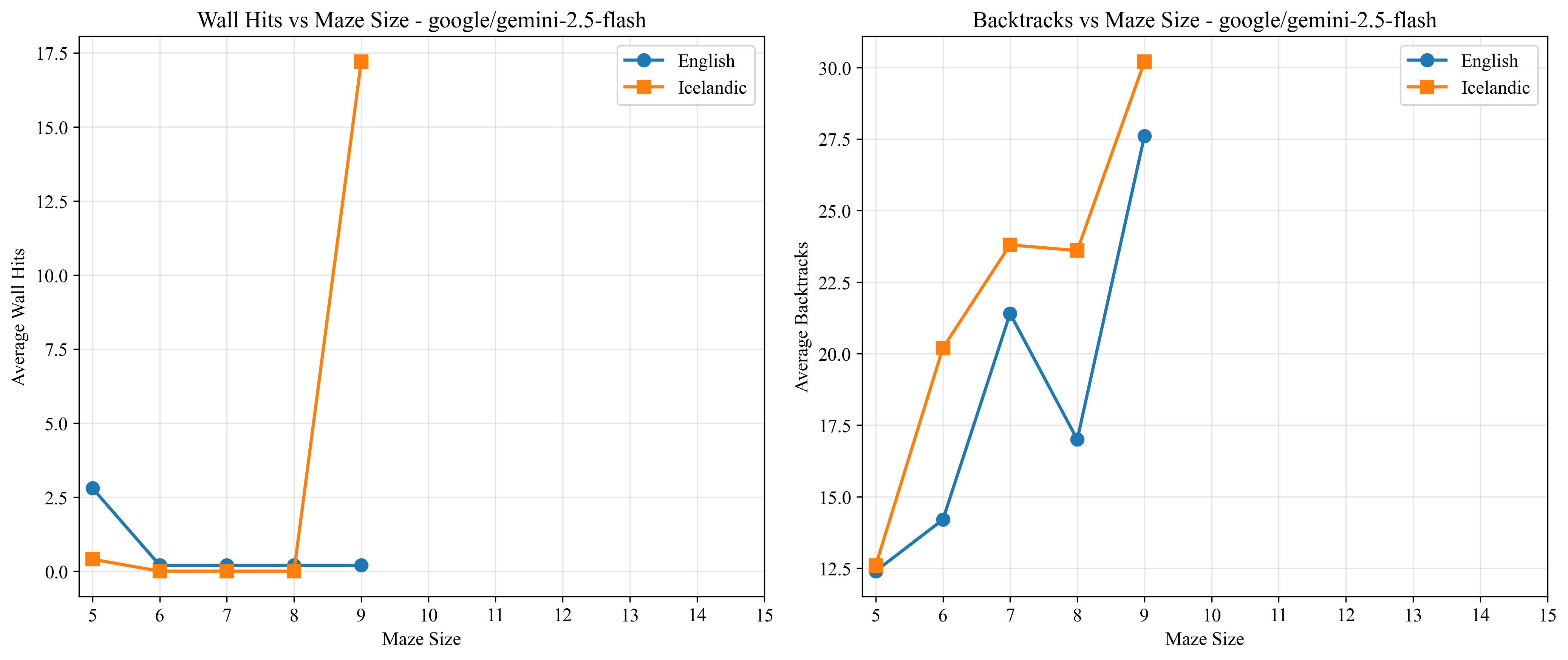}
\caption{Wall hits and backtracking patterns for Gemini 2.5 Flash across maze sizes. Note the erratic wall collision spike in Icelandic evaluations.}
\label{fig:metrics_gemini_flash}
\end{figure}

\begin{figure}[htbp]
\centering
\includegraphics[width=\columnwidth]{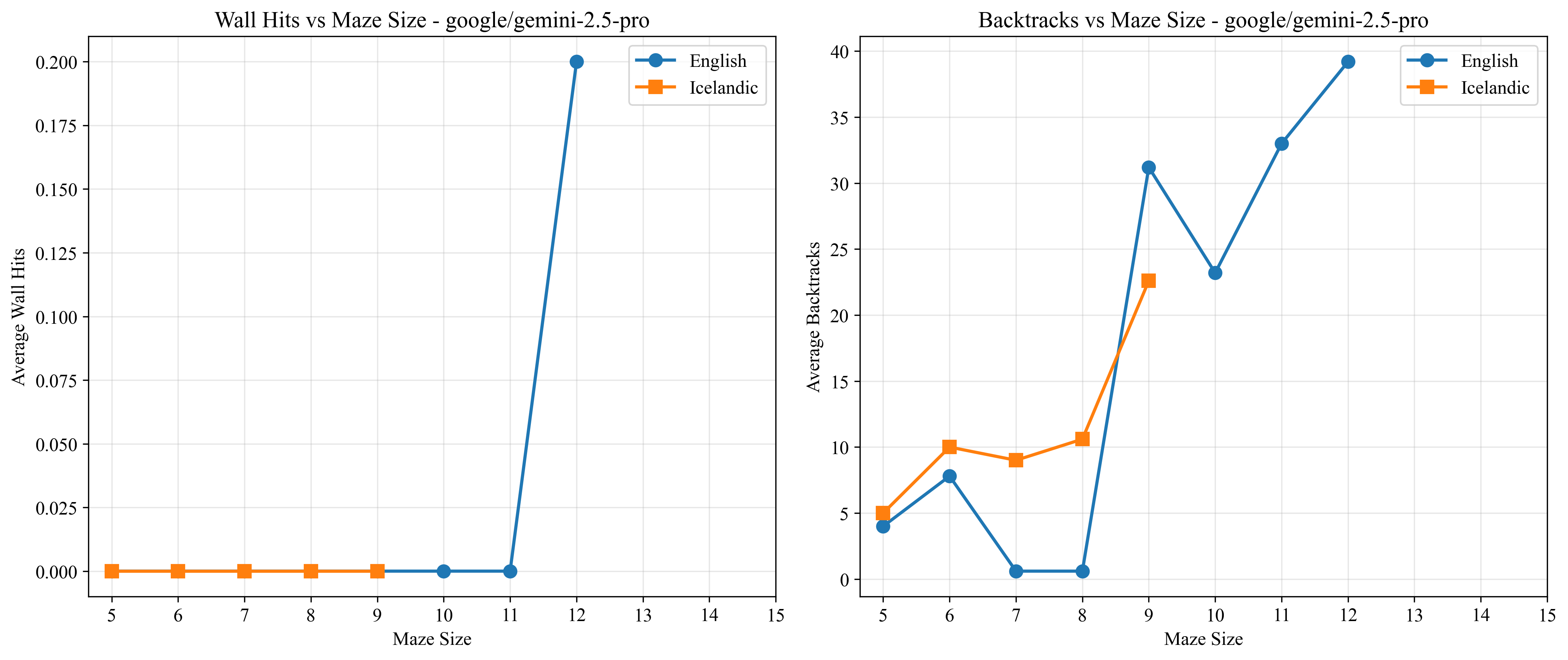}
\caption{Wall hits and backtracking patterns for Gemini 2.5 Pro across maze sizes.}
\label{fig:metrics_gemini_pro}
\end{figure}

\begin{figure}[htbp]
\centering
\includegraphics[width=\columnwidth]{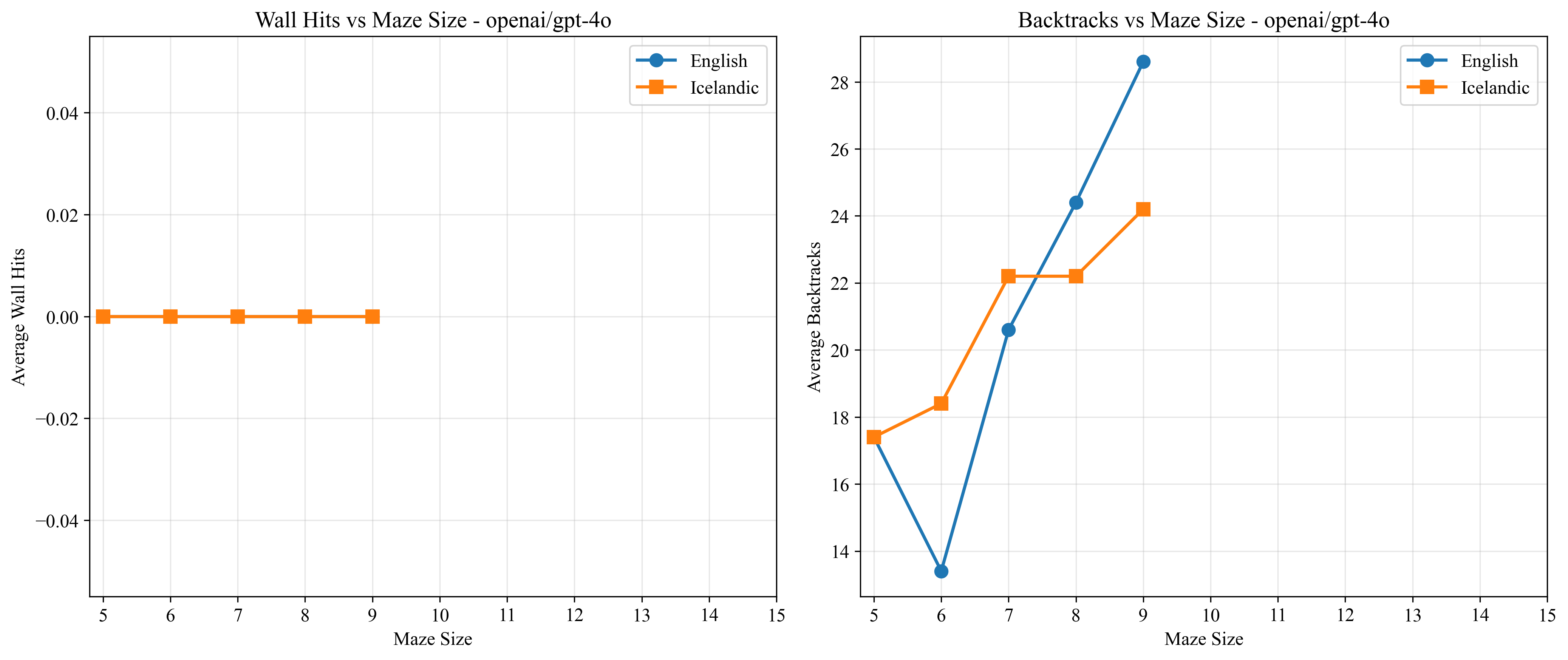}
\caption{Wall hits and backtracking patterns for GPT-4o across maze sizes.}
\label{fig:metrics_gpt4o}
\end{figure}

\begin{figure}[htbp]
\centering
\includegraphics[width=\columnwidth]{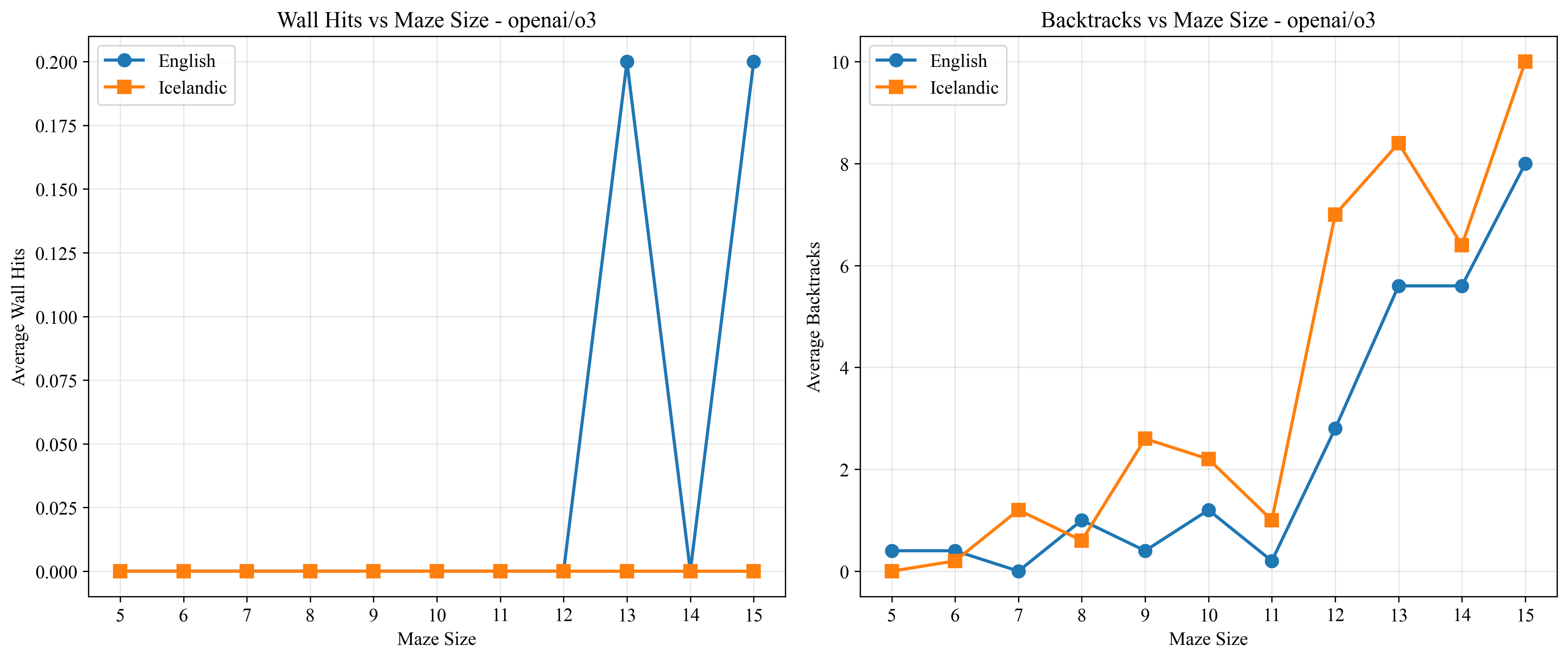}
\caption{Wall hits and backtracking patterns for OpenAI O3 across maze sizes. Note the consistently low values across all conditions, demonstrating exceptional navigation precision.}
\label{fig:metrics_o3}
\end{figure}

The analysis reveals several key patterns. Most models exhibit a steep escalation in backtracking as maze size increases, with some averaging over 80 backtracks on paths that optimally require only 20 to 30 steps. This performance degradation is exacerbated by language, as Icelandic evaluations consistently show higher backtracking rates than their English counterparts, suggesting that linguistic context impacts spatial memory maintenance. While wall collision rates are generally low, some models show instability, such as an erratic spike from Gemini 2.5 Flash in an Icelandic test. In stark contrast, O3 demonstrates qualitatively different navigation behavior, maintaining near-zero values for both metrics across all conditions. Taken together, these patterns corroborate our main findings about the fundamental limitations in current LLMs' spatial reasoning and the significant influence of language on their navigation performance.

\end{document}